\documentclass[10pt,twocolumn,letterpaper]{article}

\usepackage[pagenumbers]{cvpr}
\usepackage[export]{adjustbox}
\usepackage{graphicx}
\usepackage{floatrow}
\usepackage{subcaption}
\usepackage{array}
\usepackage{marvosym}  
\usepackage{fancyhdr}

\definecolor{cvprblue}{rgb}{0.21,0.49,0.74}
\usepackage[pagebackref,breaklinks,colorlinks,allcolors=cvprblue]{hyperref}

\newlength{\yuvionheadextra}
\fancypagestyle{yuvionheader}{%
  \fancyhf{}%
  \fancyhead[L]{\raisebox{0pt}{\includegraphics[height=17pt]{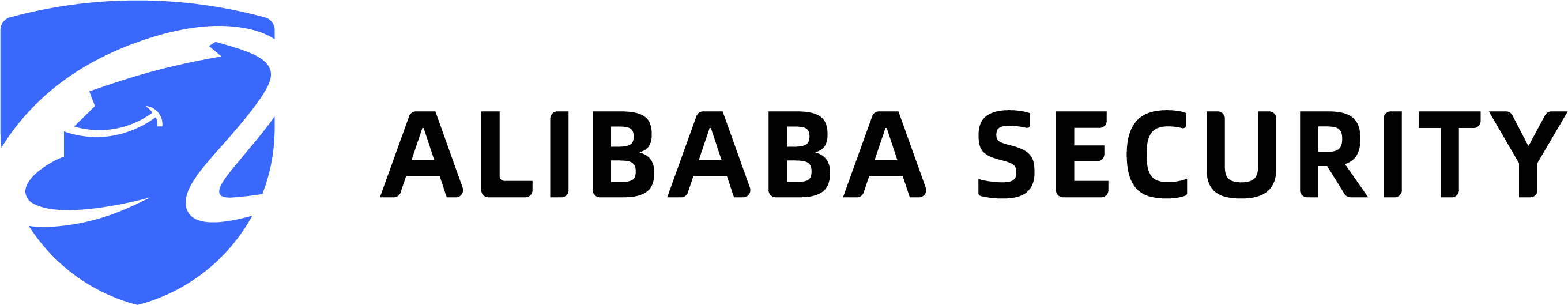}}}%
  \fancyhead[C]{}%
  \fancyhead[R]{\raisebox{0pt}{\includegraphics[height=17pt]{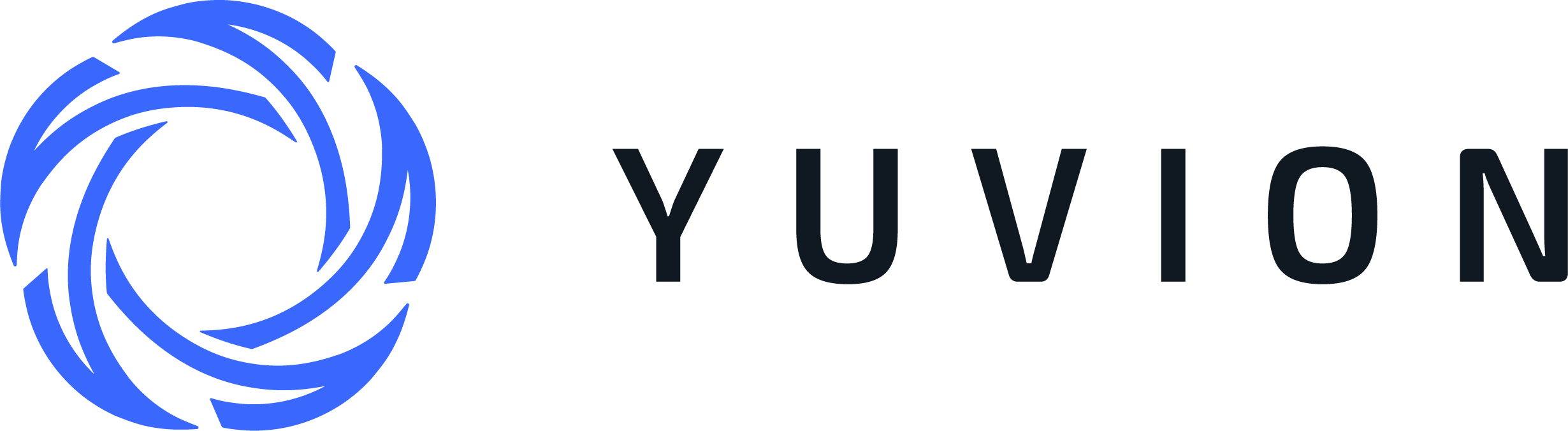}}}%
  \fancyfoot[C]{\thepage}%
}

\def\paperID{10624} 
\def\confName{CVPR}
\def\confYear{2026}

\title{SIMPLEPOSTER: A SIMPLE BASELINE FOR PRODUCT
POSTER GENERATION}

\author{
Benlei Cui$^{1*}$ \quad
Fangao Zeng$^{2*,\text{\Letter}}$ \quad
Weitao Jiang$^{2*}$ \quad
Yuwen Zhai$^{2}$ \quad
Haiwen Hong$^{1\dagger}$ \\
Longtao Huang$^{1}$ \quad
Hui Xue$^{1}$ \quad
Wenxiang Shang$^{2}$ \quad
Pipei Huang$^{2}$ \\
$^{1}$Yuvion Team, Alibaba Group \quad
$^{2}$Taobao $\&$ Tmall Group of Alibaba \\
{\tt\small \{cuibenlei.cbl,zengfangao.zfg,jiangweitao.jwt,zhaiyuwen.zyw,honghaiwen.hhw,} \\
{\tt\small kaiyang.hlt,hui.xueh,shangwenxiang.swx,pipei.hpp\}@alibaba-inc.com}
}

\begin{document}
\maketitle
\pagestyle{yuvionheader}
\thispagestyle{yuvionheader}
\begingroup
\renewcommand{\thefootnote}{\fnsymbol{footnote}}
\footnotetext{%
  $^{*}$Equal contribution. \quad
  $^{\text{\Letter}}$Corresponding authors. \quad
  $^{\dagger}$Project lead.
}
\endgroup

\begin{abstract}

Product poster generation poses distinct challenges beyond general poster design, requiring both faithful preservation of product appearance and precise control over dense, multi-line text layouts. Prior methods typically adopt inpainting frameworks augmented with auxiliary modules such as ControlNet and OCR encoders. However, these approaches introduce architectural complexity and computational overhead while still suffering from text errors and subject extension artifacts.
We present SimplePoster, a simple yet effective inpainting-based framework that achieves faithful subject preservation and accurate, position-controllable text rendering without external controllers. Our approach builds on two observations: (1) full-parameter fine-tuning of the base model effectively suppresses subject extension, outperforming ControlNet-based alternatives; and (2) a zero-cost character-level position encoding enables geometry-aware text generation without dedicated layout modules. Experiments show that SimplePoster achieves a $98.7\%$ subject preservation rate, compared to $55.2\%$ for SeedEdit 3.0 and $85.3\%$ for PosterMaker, while also improving text rendering accuracy. Code, models, benchmark and a part of training data will be available at \url{https://github.com/Alibaba-VELLDEPTH/SIMPLEPOSTER}.


\end{abstract}

\begin{figure}[!t]
  \centering
  \includegraphics[width=\textwidth]{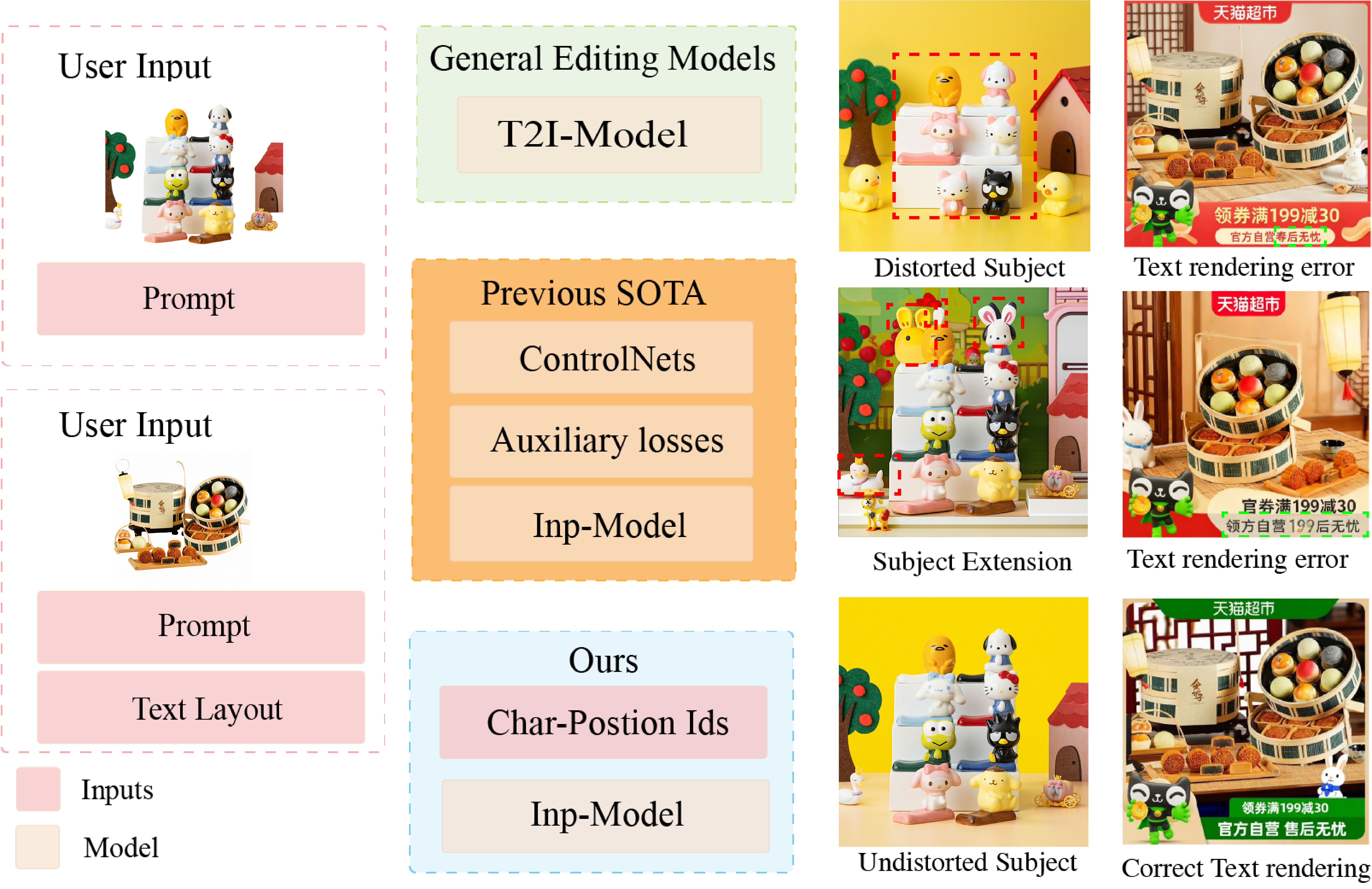}
  \caption{General editing models (top) fail to preserve subjects due to text-to-image framework. Specialized methods (middle), adopting inpainting with ControlNet, overcomes subject distortion but still exhibit subject extension (The artifact is enclosed by red dotted line boxes. \textbf{Our Solution:} SimplePoster (bottom) eliminates ControlNet dependencies while achieving strict subject preservation and precise text rendering on complex and multi-line layout.}
  \label{fig:intro-demo}
\end{figure}

\section{Introduction}
Product poster generation presents distinct challenges beyond general poster design. While general image composition systems prioritize aesthetic appeal through harmonious layouts and backgrounds, commercial applications, particularly in e-commerce, impose two critical constraints: (1) \textbf{faithful preservation} of product material properties and geometric structure, where even minor distortions may mislead consumers; and (2) precise rendering of \textbf{dense, multi-line promotional text} conveying detailed marketing information. These requirements demand not only high-fidelity subject preservation but also fine-grained spatial control over text placement.

Recent general image editing models,including FLUX-Kontext~\cite{batifol2025flux}, SeedEdit3/DreamPoster\footnotemark[1]~\cite{wang2025seededit, hu2025dreamposter}, Step1x-Edit~\cite{liu2025step1x}, Gemini 2.5 Flash~\cite{google2025gemini-2.5-flash}, and GPT-4O~\cite{gpt4o20250325}, show promise for poster generation through reference image conditioning. However, they consistently fail to meet the stringent fidelity and controllability demands. As shown in Figure~\ref{fig:intro-demo} (top) and Figure~\ref{fig:no-text-cmp}, these models frequently introduce structural deformations, texture corruption, color shifts, or text collapse. Such limitation is likely because their text-to-image frameworks lack explicit subject preservation mechanisms. Moreover, their prompt-based control cannot precisely specify multi-line text positioning.

Specialized poster generation methods~\cite{Gao_2025_CVPR, chen2025t-star, li2023planning, wang2025generate, lin2023autoposter, du2024towards} typically adopt an inpainting paradigm, keeping pre-segmented product regions fixed while synthesizing backgrounds and text. While theoretically ensuring subject preservation, these approaches still exhibit \textit{subject extension} artifacts (Figure~\ref{fig:intro-demo}, middle) and suffer from inaccurate text rendering.

\footnotetext[1]{Both SeedEdit~3.0 and DreamPoster are closed-source and appear to share the same public image generation API in Jimeng App. The internal routing mechanism of the API is not disclosed, and in practice, we use their latest available version, \texttt{SeedDream4.0}, for all evaluations.} 

To address these issues, state-of-the-art works incorporate auxiliary modules like ControlNet~\cite{zhang2023adding} for structural conditioning. The current stata-of-the-art, PosterMaker~\cite{Gao_2025_CVPR}, further introduces a subject extension detector for reinforcement learning and integrates an OCR-aware ControlNet for text rendering (as in Fig~\ref{fig:pipeline}(a)). Despite their effectiveness, these methods fail to fully eliminate subject extension while increasing architectural complexity, training overhead, and inference latency.

In this work, we try to streamline such complex pipelines with a simple framework. We present \textbf{SimplePoster}, a minimalist inpainting framework that significantly improves subject preservation and text layout control \emph{without auxiliary components} like ControlNet. Built upon the off-the-shelf FLUX-Fill~\cite{flux-fill} inpainting model, SimplePoster follows PosterMaker's task setting: given a product image and structured text specifications (content + bounding boxes), generate a compelling poster while preserving the original subject intact. Figure~\ref{fig:intro-demo} illustrates the critical distinctions between SimplePoster and existing approaches in preserving product integrity and text fidelity.

Our approach is grounded in two key insights. First, through investigation in Section~\ref{extension-exp}, we demonstrate that \textbf{full-parameter fine-tuning}, instead of  ControlNet-style pluggins, is essential for eliminating subject extension. Quantitatively, full tuning reduces the extension rate from $41\%$ (original FLUX-Fill) to $0.6\%$, outperforming ControlNet-augmented ($23.6\%$). It indicates \textbf{bridging the domain gap requires direct adaptation of the base model} rather than auxiliary modules. We attribute this to the fundamental discrepancy between standard inpainting (random small patches) and poster generation (large background regions).

Second, we propose a \textbf{Character Position Encoding} strategy. Instead of assigning all text tokens a constant $(0,0)$ coordinate as in FLUX-Fill, we bind each character token to spatial coordinates derived from its target line's bounding box. This minimal modification enables spatially grounded attention in the DiT backbone, facilitating precise text placement without glyph images, OCR features, or layout encoders. Crucially, this design also enables efficient learning of new character sets (e.g., Chinese) within a \textit{single training stage}, whereas prior methods require dedicated multi-stage training for such capabilities.

Comprehensive evaluations demonstrate SimplePoster achieves \textbf{98.7\% subject preservation rate} under strict human assessment, surpassing PosterMaker and general editing models by significant margins. It also achieves superior text accuracy in complex multi-line layouts and competitive results in prompt following and visual appeal. Our contributions are:
\begin{itemize}
    \item We identify full-parameter fine-tuning as the critical solution for subject extension, outperforming ControlNet-based approaches;
    \item We propose a novel, simple and lightweight character position encoding enabling layout-aware text generation and cross-lingual adaptation without external modules and multi-stage training. 
    \item A minimalist yet state-of-the-art baseline for product poster generation with near-perfect subject fidelity.
\end{itemize}
\section{Related Works}

\subsection{General Image Editing and Universal Image Generation}
Recent years have witnessed rapid progress in general-purpose image editing and universal generation models, including FLUX-Kontext~\cite{batifol2025flux}, Step-1X~\cite{liu2025step1x}, SeedEdit~\cite{wang2025seededit}, Gemini 2.5 Flash~\cite{google2025gemini-2.5-flash}, and GPT-4O~\cite{gpt4o20250325}. These models unify various vision tasks under text-to-image frameworks, producing high-quality outputs from natural language instructions.
Diffusion models have advanced rapidly in image generation, with recent progress spanning controllable image editing, virtual try-on, generation quality prediction, and efficient sampling~\cite{sun2025attentive,liu2025erasediffusion,li2025shoefit,cui2026diffusionprobe,cui2026tcpade}.

For commercial-use product poster generation, they face two fundamental limitations. First, they lack mechanisms to enforce strict subject preservation, often introducing distortions in shape, texture, or color of the input product. Second, their layout control relies solely on text prompts, which offer limited precision for specifying positions of multiple text lines in complex arrangements. As a result, they fall short in meeting the rigorous demands of commercial design.

\subsection{Automatic Poster Generation}

Prior work on automatic poster generation falls into two categories: artistic posters and product-specific posters.
\paragraph{Artistic Poster Generation.}
Methods such as Posta, PosterCraft and DesignDiffusion~\cite{chen2025posta, chen2025postercraft, wang2025designdiffusion} focus on creative layout design, visual harmony, and stylistic rendering. They often generate abstract or illustrative scenes without user-provided subjects, prioritizing aesthetics over fidelity. Consequently, these approaches are not suitable for product-centric applications where subject integrity is paramount.
\paragraph{Product Poster Generation.}
Early approaches~\cite{li2023planning, wang2024prompt2poster, chen2024anyscene, chen2025t-star, lin2023autoposter, wang2025generate} adopt multi-stage pipelines involving separate modules for layout planning, attribute prediction, and image synthesis. Most leverage inpainting-based diffusion models to preserve the product subject, but rely heavily on ControlNet~\cite{zhang2023adding} for spatial conditioning of layout and text.
The current state-of-the-art, PosterMaker~\cite{Gao_2025_CVPR}, unifies the pipeline into a single stage and incorporates techniques from visual text generation techniques~\cite{yang2023glyphcontrol, tuo2023anytext, liu2024glyph-byt5}: injecting fine-grained character-level OCR features to the diffusion model to improve text rendering. It further train a subject extension detector as a reward model to suppress subject extension. While effective, these enhancements come at the cost of increased system complexity and reduced reproducibility.
In contrast, our work demonstrates that neither controller such as ControlNet nor auxiliary reward models are necessary. With careful model adaptation and a simple positional encoding strategy, we achieve superior performance using a minimalist architecture.

\section{How to eliminate subject extension?}
\label{extension-exp}

We investigate the prevalent subject extension problem in inpainting-based poster generation and analyze mitigation strategies. Using the state-of-the-art FLUX-Fill inpainting model as baseline, we observe a $41\%$ subject extension rate on our benchmark dataset (Table~\ref{tab:exp-extension}). Following prior work~\cite{Gao_2025_CVPR}, we integrate and tune ControlNet~\cite{zhang2023adding} for structural conditioning while freezing the base DiT backbone. This reduces extension to $23.6\%$, indicating limited improvement.

We question whether this stems from a domain gap: standard inpainting models are typically trained on datasets where small, randomly sampled patches are masked, while poster generation requires synthesizing large background regions adjacent to fixed products. Merely adding controllers cannot bridge this gap—direct adaptation of the base model is required. To validate this, we try to tune the model with Low-Rank Adaptation (LoRA)~\cite{hu2022lora}. Results reveal that LoRA-despite fewer parameters—reduces extension to 2.8\%, significantly outperforming ControlNet-augmented frozen model. Crucially, we further full-parameter tune the basemodel,  and reduce the extension to 0.6\%, demonstrating near-complete mitigation (Table~\ref{tab:exp-extension}). 

These findings establish that direct model adaptation, particularly full-parameter tuning, fundamentally outperforms auxiliary modules for structural control. Rather than adding complexity to misaligned models, refining internal representations addresses the root cause of subject extension. Visualizations of the subject extension cases are provided in the \textit{\color{red}Supplementary Materials}.

\begin{table}
  \begin{adjustbox}{width=1.0\linewidth}
  \begin{tabular}{c | c}
    \hline
    & Subject Extension Rate $\downarrow$ \\
    \hline
    baseline & $41\%$\\
    ControlNet & $23.6\%$ \\
    Lora Tuning & $2.8\%$\\
    Full-Parameter Tuning & $0.6\%$\\
    \hline
  \end{tabular}
  \end{adjustbox}
  \caption{quantitative evaluation of different approaches on subject extension rate.}
  \vspace{-8pt}
  \label{tab:exp-extension}
\end{table}

\begin{figure*}[!t]
  \centering
\includegraphics[width=\textwidth]{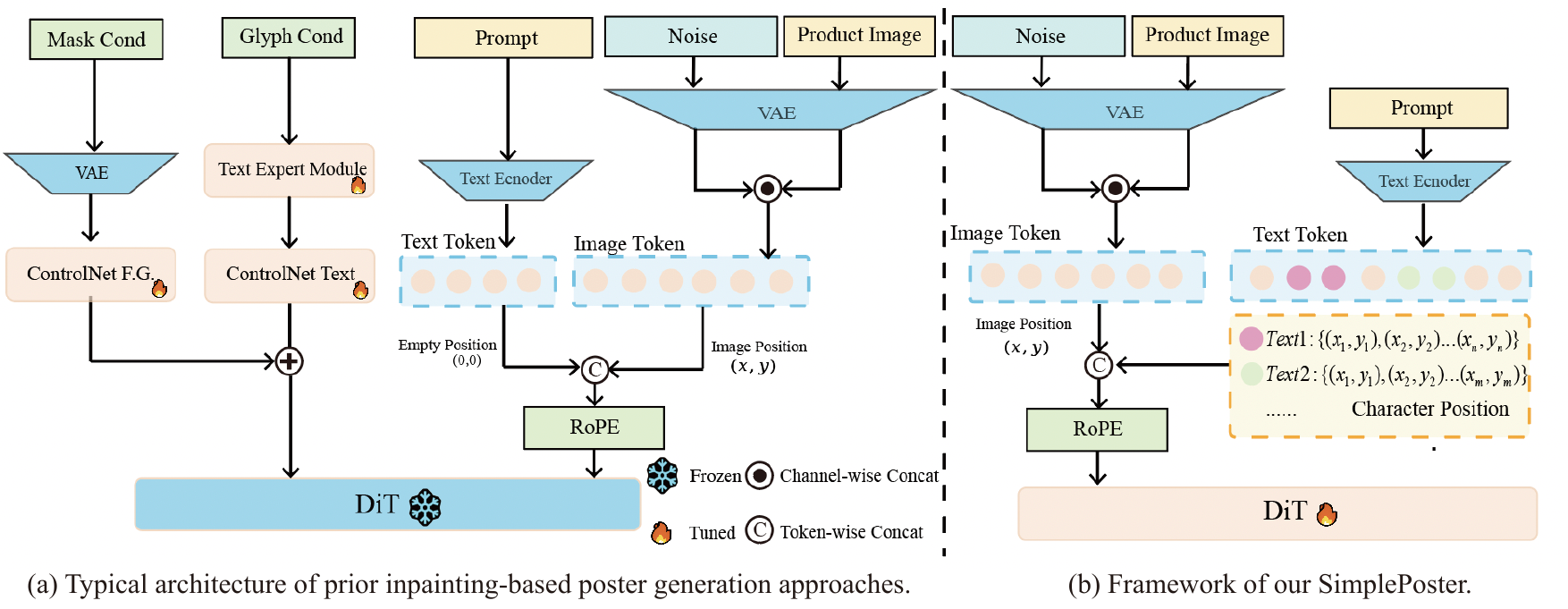}
  \caption{Architectural comparison between prior inpainting-based frameworks and our SimplePoster. (a) Prior works rely on auxiliary ControlNets for structural conditioning and OCR encoding, increasing architectural complexity. (b) Our SimplePoster eliminates all external controllers through full-parameter fine-tuning and Character Position Encoding.}
  \label{fig:pipeline}
\end{figure*} 

\section{Method}
\subsection{Task Formulation}
We adopt the task formulation from PosterMaker~\cite{Gao_2025_CVPR}: given an input triplet $(I, \mathcal{P}, \mathcal{B})$ where:
\begin{itemize}
    \item \textbf{Product image} ($I$): User-provided product image with white background. Non-conforming inputs are preprocessed using segmentation and matting to isolate the product on white background.
    
    \item \textbf{Text line positions} ($\mathcal{B}$): Set of bounding boxes $b_i = (x_l, y_t, x_r, y_b)$ specifying spatial locations for each text line.
    
    \item \textbf{Prompt} ($\mathcal{P}$): Natural language description of background scene and text content, augmented with coarse layout cues (e.g., "Text 'Fast delivery' at top center").
\end{itemize}

Given this input triplet, the model generates a photorealistic product poster that adheres to the prompt, accurately renders each text line within its specified region, and preserves the original product subject. Formally, the goal is to learn a generation function $G(I, \mathcal{P}, \mathcal{B}) \to I^*$ such that $I^*$ satisfies all spatial, semantic, and fidelity constraints.

\subsection{Framework}
We build SimplePoster upon the state-of-the-art inpainting model FLUX-Fill~\cite{flux-fill}, which consists of a VAE, a Diffusion Transformer (DiT)~\cite{dit}, and a T5 text encoder~\cite{raffel2020t5}. While preserving the overall simplicity of the pipeline, we introduce three key modifications to adapt it for product poster generation:

(1) \textbf{Multilingual prompt support}: Since T5 only accepts English input, we replace it with Qwen2.5-VL~\cite{bai2025qwen2}, a strong vision-language model supporting both Chinese and English prompts, following Step1x-Edit~\cite{liu2025step1x}.

(2) \textbf{Full-parameter fine-tuning}: Unlike prior works that freeze the base model and train auxiliary controllers (e.g., ControlNet), we end-to-end fine-tune the entire DiT backbone. As Section~\ref{extension-exp} demonstrates, this eliminates subject extension without external modules.

(3) \textbf{Character-level position encoding for precise layout control}: We propose a simple yet effective spatial grounding mechanism that assigns each character token a meaningful 2D coordinate derived from its target text line’s bounding box. These coordinates are encoded via RoPE~\cite{su2024roformer} for attention computation in the DiT, enabling accurate text placement, without requiring glyph images, OCR features, or external layout models. This design introduces no architectural changes and thus incurs no inference overhead.

Figure~\ref{fig:pipeline}(b) illustrates the framework. Compared to the typical inpainting-based pipeline of prior art ((Figure~\ref{fig:pipeline}.(a)), such as PosterMaker~\cite{Gao_2025_CVPR}, our framework is significantly simpler, as it eliminates the need for auxiliary conditioning modules like ControlNet.

\subsection{Character Position Encoding Strategy}
\label{character-position}
In the original FLUX-Fill framework, RoPE~\cite{su2024roformer} converts the spatial coordinate 
$(x,y)$ of each image token into position-aware embeddings for attention computation. However, for text tokens, including those corresponding to characters to be rendered in the image, all are assigned a fixed 2D position $(0,0)$, regardless of their intended layout. This uniform positioning limits the model's ability to ground text generation in specific spatial regions.

In contrast, our method assigns each character token a spatially meaningful 2D coordinate $(x_c, y_c)$ based on its target rendering location. Specifically, given a text line consisting of $n$ characters and its bounding box $(x_l,y_t,x_r,y_b)$, we divide the box horizontally into $n$ equal sub-regions. Assuming left-to-right writing order, we assign the center of the $i$-th sub-region as the spatial position for the $i$-th character token. 

Formally, the spatial coordinate assigned to the $i$-th character is:

\begin{equation}
(x_c^i, y_c^i)=\Big(x_l+\frac{i-0.5}{n}(x_r-x_l), \frac{y_t+y_b}{2}\Big)    
\end{equation}
for horizontal left-to-right text. Vertical arrangements follow analogously.

This minimal yet effective modification enables the DiT to attend to spatially grounded text representations, thereby guiding each character to be synthesized at its user-specified location, without requiring any additional layout controllers.

\subsection{Training Protocol}
\textbf{Losses.} We adopt the standard flow matching objective used in SD3~\cite{sd3} and FLUX~\cite{flux-fill}, without introducing any auxiliary losses. 

\noindent \textbf{End-to-End Training}
Unlike prior works that rely on multi-stage training pipelines, such as separate stages for text generation and background synthesis, we employ single-stage end-to-end fine-tuning of the pretrained DiT backbone. We found that model converges fast and steadily during the training, so there is no need for multi-stage training. We attribute this to our Character Position Encoding and elaborate its impact in Section~\ref{PE_discussion}.

\begin{figure*}[!t]
  \centering
  \includegraphics[width=0.95\textwidth]{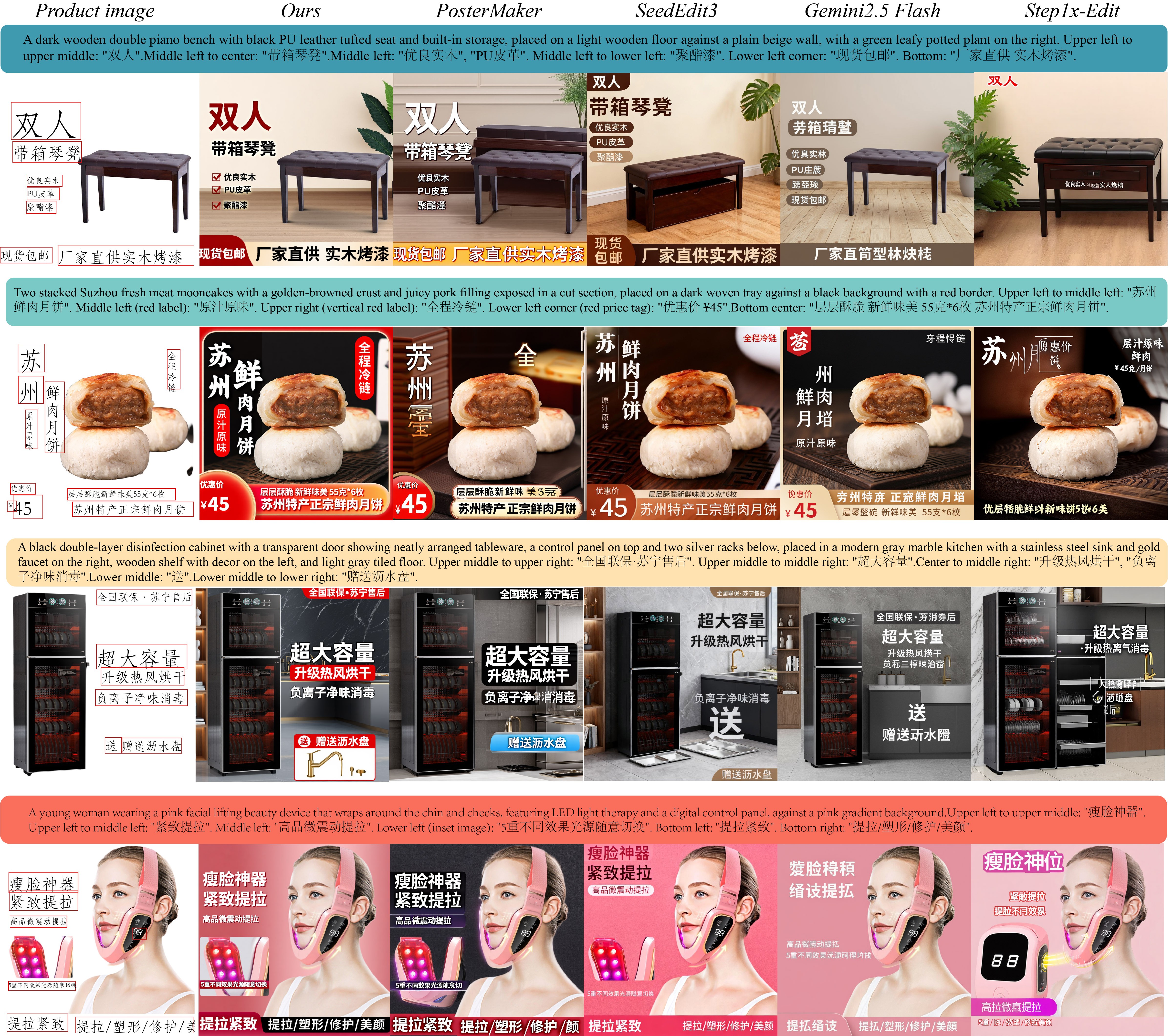}
  \caption{Qualitative comparison of promotional text rendering. Zoom for detail. Text boxes and contents are rendered at the product image for better visualization. Due to space constraints, the complete prompts are provided in the Supplementary Materials, with an abbreviated version shown in this figure}
  \label{ig:text-cmp}
\end{figure*}

\section{Experiments}
We construct a dedicated training set to train SimplePoster and establish a comprehensive benchmark for evaluating our method, SOTA general image editing models, and prior specialized approaches.

\subsection{Training Data}
We collect approximately 1.5 million product images from real-world e-commerce platforms, covering diverse categories such as shoes, toys, bags, cosmetics, and furniture. Nearly 300k of these images contain Chinese promotional text. Each training sample is constructed as a triplet $(I, \mathcal{P}, \mathcal{B})$:
(1) \textbf{Input image}: We apply a subject segmentation model followed by a matting model to extract the product subject and place it on a pure white background;  
(2) \textbf{Text line position}: We use an OCR engine to detect all text lines, their bounding boxes, and corresponding text content in the original image. Text printed directly on the product (e.g., logos or labels) is excluded to avoid conflicts with generated promotional text;  
(3) \textbf{Prompt}: We leverage Qwen2.5-VL-72B~\cite{bai2025qwen2} to generate a descriptive caption for each original image, serving as the input text prompt. Additionally, we incorporate coarse spatial descriptions of each text line into the prompt (e.g., "The text ‘In stock’ spans from the top-left corner to the top region of the image.").

\subsection{Evaluation Benchmark.}
Our test set comprises 500 carefully curated product images spanning common e-commerce categories. Notably, 200 images feature dense, multi-line Chinese promotional text—scenarios that expose the limitations of methods optimized for sparse English layouts. Each sample follows the same preprocessing pipeline as training data.

\noindent\textbf{Baseline Configuration.} General image editing models (SeedEdit~3/DreamPoster, Gemini~2.5 Flash, FLUX-Kontext, Step1x-Edit) receive only natural language prompts, as they lack support for explicit bounding box conditioning. For PosterMaker~\cite{Gao_2025_CVPR}, we omit coarse spatial descriptors from $\mathcal{P}$ since it explicitly conditions on $\mathcal{B}$; we empirically observe that providing both degrades prompt adherence.

\noindent\textbf{Inference Protocol.} All models generate at $1024 \times 1024$ resolution. General editing models receive a task prefix \texttt{Background generation task.} to indicate the task. FLUX-Kontext prompts are translated from Chinese to English due to its monolingual limitation.

\subsection{Evaluation Metrics.}
We assess performance across four critical dimensions:

\noindent\textbf{Subject Preservation Rate (SPR):} The percentage of generated images that strictly maintain product structure, texture, color, and branding, as judged by human evaluators. A sample is preserved only if zero geometric deformation, material alteration, or branding distortion is present.
    
\noindent\textbf{Text Rendering Accuracy:} Following~\cite{Gao_2025_CVPR, tuo2023anytext}, we report Sentence Accuracy (Sen. Acc) and Normalized Edit Distance (NED). We employ an off-the-shelf OCR system used in ~\cite{tuo2023anytext} to extract text from generated images and align predictions with ground truth. Sen. Acc requires exact line-level matching; NED computes the normalized Levenshtein distance, providing partial credit for near-misses.
    
\noindent \textbf{Prompt Following:} To assess adherence to the input prompt—including scene composition, stylistic guidance (e.g., “luxury style”), and semantic instructions, we conduct a user study in which 5 experienced annotators rate the degree of compliance on a 5-point Likert scale (from 1: poor to 5: excellent).
    
\noindent \textbf{Visual Appeal:} In the same user study, annotators independently rate the aesthetic quality and overall design harmony using the same 5-point Likert scale. Ratings consider factors such as layout balance, color coordination, typographic clarity, and visual attractiveness.

\subsection{Implementation Details}
For the main result in table~\ref{tab:sys-cmp-our}, we train the FLUX-Fill DiT backbone on 128 NVIDIA H20 GPUs (total batch size 512) for 3 epochs of our whole training set of 1.5M images, costing ~40 hours. We use AdamW optimizer with learning rate 5e-5 and weight decay 1e-2. Both training and inference operate at $1024 \times 1024$ resolution.

For experiments in Sec~\ref{extension-exp}, we use 32 GPUs with total batch size 128. To isolate the subject extension analysis from text rendering, we sample a subset of 300k images without promotional text.
We still train the subset for 3 epochs. For the LoRA variant, we set the rank to 64. For the ControlNet variant, we use Canny edge maps of product images as the conditioning signal. The ControlNet implementation builds upon the x-flux codebase\footnote{\url{https://github.com/XLabs-AI/x-flux}}.
\begin{table*}
  \centering
  \begin{adjustbox}{width=\linewidth}
  \begin{tabular}{c | c c c c c}
    \hline
    & Subject Preservation Rate $\uparrow$  & Sen.ACC$\uparrow$ & NED$\uparrow$ & Prompt Following$\uparrow$ & Visual Appeal$\uparrow$ \\
    \hline
    FLUX-Kontext (pro)\footnotemark[2] & $36.27\%$ & 0.076 & 0.1460 & 3.02 & 3.21\\
    Step1x-Edit & $28.8\%$ & 0.094 & 0.313 & 2.55 & 3.08\\
    Gemini2.5Flash & $51.4\%$ & 0.3231 & 0.6303 & \textbf{3.82} & 4.12\\
    SeedEdit3.0/DreamPoster\footnotemark[1] & $55.2\%$ & 0.6434 & 0.7825 & 3.73 & \textbf{4.22} \\
    PosterMaker & 85.3\% & 0.5757 & 0.7392 & 3.32 & 3.74\\
    Ours & \textbf{98.7\%}  & \textbf{0.7133}  & \textbf{0.8062} & 3.55 & 4.03 \\
    \hline
  \end{tabular}
  \end{adjustbox}
  \caption{Quantitative evaluation of different approaches on our benchmark. The best results are marked in bold.}
  \vspace{-8pt}
  \label{tab:sys-cmp-our}
\end{table*}

\begin{table}
\begin{tabular}{c|cc}
\hline
 & NED$\uparrow$ & Sen.Acc$\uparrow$ \\
\hline
Full-setting & \textbf{0.8062} & \textbf{0.7133} \\
Full setting w/o Character PE & 0.5484 & 0.2494 \\
\hline
\end{tabular}
\caption{Ablation on Character Position Encoding.}
\label{tab:char-pe}
\end{table}



\subsection{Main Results}
\textbf{Comparison with State-of-the-Art.}
Table~\ref{tab:sys-cmp-our} presents comprehensive quantitative comparisons. SimplePoster achieves a \textbf{98.7\% Subject Preservation Rate}, substantially surpassing the best general editing model SeedEdit~3/DreamPoster ($55.2\%$) and the specialized PosterMaker ($85.3\%$). In text rendering, SimplePoster attains \textbf{71.33\% Sen. Acc} and \textbf{0.8062 NED}, outperforming all baselines including PosterMaker (57.57\% Sen. Acc, 0.7392 NED).

\footnotetext[2]{Since FLUX-Kontext is unable to generate Chinese characters, its low text accuracy in our benchmark is expected.}

\begin{figure*}[!t]
  \centering
\includegraphics[width=\textwidth]{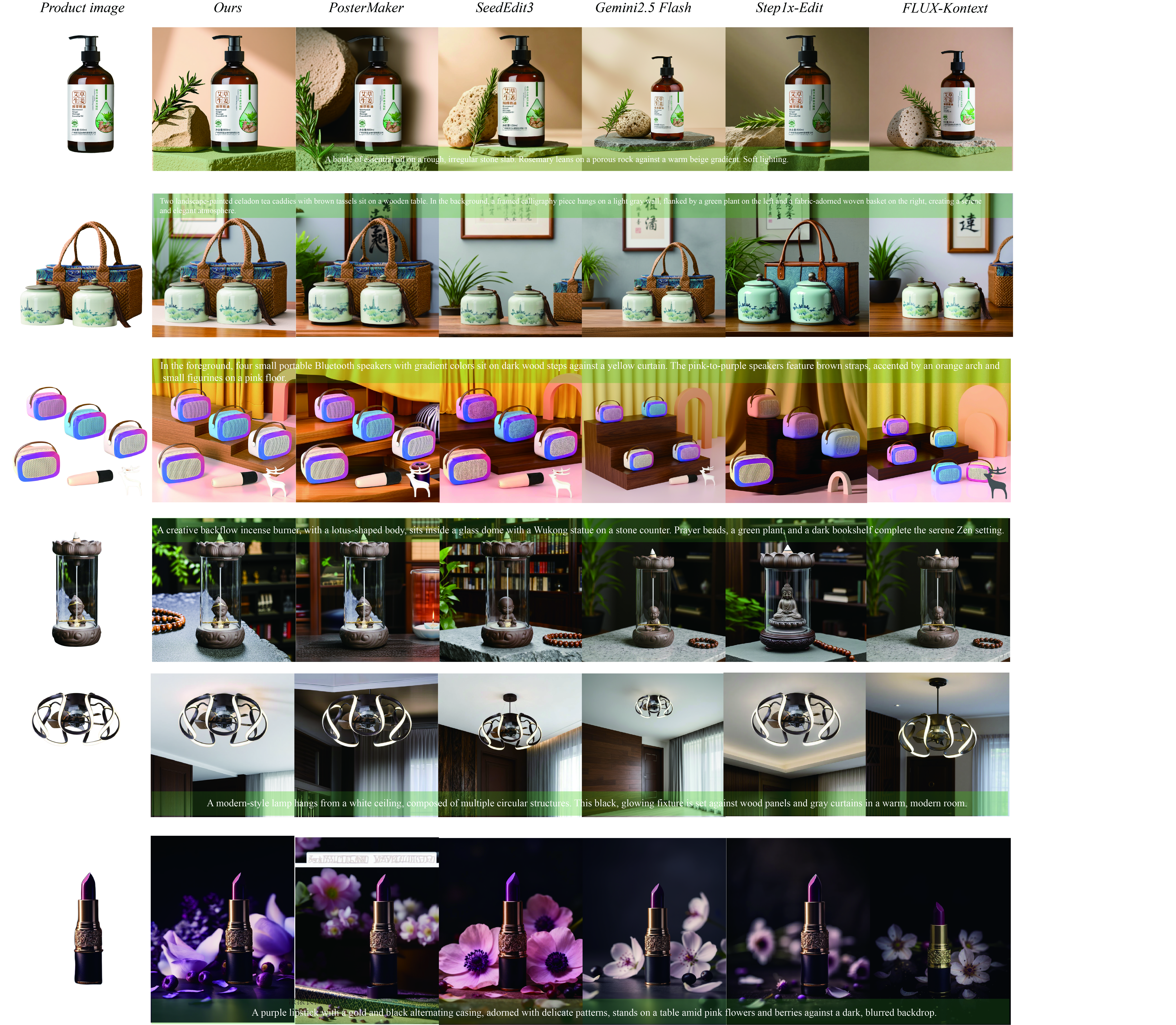}
  \caption{Qualitative comparison on samples without promotional texts. Best viewed on screen when zoomed in to observe fine-grained textures and small text details. Due to space constraints, the complete prompts are provided in the Supplementary Materials, with an abbreviated version shown in this figure}
  \label{fig:no-text-cmp}
\end{figure*}

\noindent\textbf{Analysis of Subject Preservation Failures.} We notice the SOTA editing models often fail to maintain strict subject preservation. As shown in Table~\ref{tab:sys-cmp-our}, the best editing model SeedEdit3.0/DreamPoster achieves only $55.2\%$ subject preservation rate. We analyze their failure cases and identify three major types of artifacts, illustrated in Figure~\ref{fig:no-text-cmp}
\begin{itemize}

\item \textbf{Collapse of high-frequency textures and text}: This is the most prevalent failure mode. Whenever textual labels or patterns of high-frequency are present on the product, editing models frequently distort, blur, or erase these fine details. The essential oil bottle in the first row exemplifies this issue: all editing models fail to preserve the printed text, leading to illegible or altered branding.

\item \textbf{Structural changes}: Editing models tend to alter the geometric structure of products with complex shapes. In the second row, both Step1x-Edit and PosterMaker add a base to the teapot. In the fourth row, all editing models except Gemini2.5 Flash change the Wukong sculpture, indicating unintended semantic edits. In the fifth row, SeedEdit and FLUX-Kontext extend the flush-mount light with an extra rod segment, transforming it into a hanging lamp.

\item \textbf{Material and color alterations}: As shown in the second row, Step1x-Edit alters the color of the teapot.
\end{itemize}

Even the current SOTA specialized method PosterMaker, which employs ControlNet and a reward model to suppress subject extension, achieves an excellent subject preservation rate of $85.3\%$, but still suffers from unintended subject extension. For instance, in Figure~\ref{fig:no-text-cmp}, PosterMaker extends the oil bottle in the first row and the bowl in the second row.

In contrast, SimplePoster achieves near-perfect preservation rate of $98.7\%$ , faithfully maintaining the product’s structure, material properties, and texture without any geometric or semantic distortions. It significantly outperforms PosterMaker and SOTA editing models.

This superior performance highlights the effectiveness of the inpainting-based framework, once the subject extension issue is properly addressed through full-parameter fine-tuning, over T2I-based general editing models, which lack explicit mechanisms for subject fidelity control.

\noindent\textbf{Analysis of Text Rendering.}
\label{discuss:text-performance}
Since FLUX-Kontext cannot generate Chinese characters, it is excluded from this comparison. As shown in Figure~\ref{ig:text-cmp}, on multi-line and complex layout cases, Gemini 2.5 Flash and Step1x-Edit produce numerous character-level errors, including incorrect glyphs, missing text, and misaligned lines. SeedEdit3 performs significantly better than these models: the generated text lines are generally correct, though some characters appear blurred or slightly distorted.

Quantitatively, as reported in Table~\ref{tab:sys-cmp-our}, even after training on Chinese data, Step1x-Edit achieves only $0.094$ Sen. Acc and $0.313$ NED. Gemini 2.5 Flash performs better but still lags behind the specialized model PosterMaker (0.5757 Sen. Acc, 0.7392 NED). This is expected, as these general editing models rely solely on coarse natural language prompts without access to precise text line coordinates, making accurate spatial layout control inherently challenging.

Notably, SeedEdit3.0/DreamPoster achieves remarkably strong text rendering performance despite using only prompt-based spatial cues. It surpasses not only other general editing models but also the specialized PosterMaker. We attribute this advantage primarily to its underlying base model, SeedDream4.0~\cite{seedream2025seedream}, which may have been trained on large-scale multilingual and layout-aware data, endowing it with superior Chinese text synthesis capabilities compared to other models.

SimplePoster achieves the best overall text rendering performance, surpassing all baselines by a clear margin. This demonstrates that our character-level position encoding strategy, combined with full-parameter fine-tuning, enables highly accurate and spatially controllable text synthesis, without relying on external controllers or OCR-derived features.

\noindent\textbf{Trade-offs in Visual Quality.}
As shown in Table~\ref{tab:sys-cmp-our}, SimplePoster quantitatively outperforms PosterMaker in both prompt following and visual appeal, but lags behind Gemini 2.5 Flash and SeedEdit3.0/DreamPoster. The last row of Figure~\ref{fig:no-text-cmp} presents a representative failure case of ours: the color of the petals is incorrectly rendered, the purple stamens are missing despite being specified in the prompt, and the generated berries appear unrealistic. This performance gap can be partly attributed to the nature of our training data. Since the images are sourced from real-world e-commerce platforms, they often prioritize product clarity over background diversity and aesthetic quality, resulting in less visually rich or stylistically varied compositions.

\subsection{Ablation Studies}
\label{sec:ablation}

\noindent\textbf{Character Position Encoding.} 
We validate the necessity of our Character Position Encoding for precise text rendering and efficient cross-lingual adaptation. As shown in Table~\ref{tab:char-pe}, removing the Character PE causes catastrophic degradation: Sen. Acc plummets from \textbf{71.33\% to 24.94\%} and NED from \textbf{0.8062 to 0.5484}. Notably, without explicit coordinate supervision, the model fails to converge on Chinese character generation within 3 epochs, whereas CPE enables simultaneous layout control and multilingual learning without multi-stage training. 

\noindent\textbf{Data Efficiency.} 
While Section~\ref{extension-exp} demonstrates that full-parameter tuning eliminates subject extension, we further investigate the minimal data requirements for this capability. Surprisingly, as detailed in the \textit{\color{red}Supplementary Materials}, merely 3K training images suffice to reduce the extension rate from $41\%$ to $3.6\%$, outperforming ControlNet trained on 300K images. 

\section{Generalization of Character PE}
\label{PE_discussion}
Beyond product poster generation, we empirically find that Character Position Encoding enhances text rendering in general text-to-image task and generalizes to more languages. Specifically, it yields substantial improvements for Korean and Japanese, particularly in multi-line layouts.  Moreover, it accelerates training convergence. Standard text-to-image models rely on text-only prompts containing coarse spatial descriptions (e.g., ``text at top center''), which introduce ambiguity: multiple spatial configurations satisfy the same prompt, slowing the learning. By providing precise per-character coordinates, our Character Position Encoding eliminates this spatial ambiguity. Details of the text-to-image experiment are provided in the \textit{\color{red}Supplementary Materials}.

\section{Limitations}
While SimplePoster achieves strong performance in product poster generation, several limitations remain: dependency on segmentation and matting quality, inability to modify product attributes under inpainting constraints. We provide detailed discussion in the \textit{\color{red}Supplementary Materials}.

\section{Conclusion}
In this work, we address product poster generation, where subject fidelity and text layout control are crucial for commercial deployment. We identify critical failure modes in existing approaches: structural distortions in general-purpose models and subject extension in specialized methods.
Our key insight is that full-parameter fine-tuning alone,without external controllers, eliminates subject extension and achieves near-perfect preservation. Furthermore, we introduce character position encoding that grounds each token in spatial layout, enabling precise text synthesis without multi-stage training or OCR-based control.
Combining these, we propose \textbf{SimplePoster}, achieving state-of-the-art performance with remarkable architectural simplicity. Experiments demonstrate superior subject preservation and text accuracy over all prior methods, establishing a new strong baseline.



\clearpage
\setcounter{page}{1}

\def\paperID{10624} 
\def\confName{CVPR}
\def\confYear{2026}


\section*{Supplementary Material}
\appendix

This document provides: (i) detailed discussion of limitations (Section~\ref{limitation}); (ii) generalization experiments of Character Position Encoding (CPE) on multilingual text-to-image synthesis (Section~\ref{exp:t2i}); (iii) qualitative comparison of subject extension mitigation strategies (Section~\ref{fig:controlnet}); (iv) data efficiency analysis investigating minimal data requirements for eliminating subject extension (Section~\ref{supp:data-scale}); (v) automatic evaluation metrics for subject preservation using DinoV3 (Section~\ref{supp:auto-eval}); and (vi) complete text prompts for visualized examples in the main paper (Section~\ref{supp:prompts}).

\paragraph{Related team technical report.}
Yuvion VL, developed by our team, presents a multimodal foundation model for adversarial content understanding and AI safety~\cite{qiu2026yuvionvl}.


\section{Limitations}
\label{limitation}
While SimplePoster achieves strong performance in product poster generation, several limitations remain, falling into three main categories:

\noindent\textbf{Dependency on segmentation and matting quality.}
Our framework assumes the input product image has a clean white background. When this is not the case, we rely on off-the-shelf segmentation and matting models to extract the foreground. However, inaccurate extraction—such as over-segmentation (including non-product regions) or under-segmentation (truncating parts of the product)—can degrade generation quality. This issue is particularly pronounced when the original image contains dense promotional text, which may confuse the extractor and lead to incorrect masks. In such cases, the inpainting model must either reconstruct missing product content or suppress conflicting background elements, potentially compromising structural fidelity. As demonstrated in the last row of Figure 3 of the main paper, under-segmentation forces the model to complete missing sections. In contrast, general image editing models do not require explicit foreground masking and can implicitly localize the subject from context, though they risk misidentification as well.

\noindent\textbf{Inability to modify product attributes under inpainting constraints.}  
The inpainting paradigm enforces strict preservation of the unmasked product region, preventing changes to intrinsic product states. For example, if the input shows a water-filled bottle but the prompt requests an empty one, SimplePoster cannot fulfill this instruction without modifying the masked area. It is a fundamental constraint of the inpainting framework. While this design ensures high subject fidelity, it limits applicability to tasks requiring semantic edits to the product itself, such as state changes, color swaps, or style transfers within the product region.

\noindent\textbf{Text generation accuracy can be further improved.}  
 Despite leveraging precise text line coordinate, SimplePoster only slightly outperforms SeedEdit3.0 which relies solely on coarse spatial cues, in overall text rendering quality. As discussed in Section 5.5 of the main paper, we attribute this primarily to differences in base model capabilities. This suggests that while our layout control mechanism is effective, further improvements in text fidelity will depend on stronger pretraining in multilingual data.

\section{Generalization Verification of Character Position Encoding}
\label{exp:t2i}
To validate the generalization capability of our Character Position Encoding (CPE) on general text-to-image synthesis and multilingual scenarios, we conduct a pilot experiment extending FLUX.1-dev—originally an English-only model—to generate Japanese and Korean text.
\subsection{Dataset} We collect approximately 40k images containing Japanese or Korean text from social media platforms, evenly split between the two languages. Unlike product posters, these images feature relatively simple layouts, predominantly single-line text. We construct two validation sets: (i) \textit{Single-line:} images containing one text line; (ii) \textit{Multi-line:} images containing two or more text lines.
\subsection{Implementation.} We first replace the T5 text encoder in FLUX.1-dev with Qwen2.5-VL, following the main paper. We establish a baseline by fine-tuning the DiT with LoRA (rank=256) for 30 epochs. Subsequently, we incorporate our Character Position Encoding during training, applying it with $50\%$ probability per image. During inference, we disable CPE, reverting the model to standard text-to-image generation.
\subsection{Results} As shown in Table~\ref{tab:char-pe}, CPE yields substantial improvements on the multi-line validation set and moderate gains on the single-line set. Notably, multi-line text accuracy (Sen. Acc) improves by $458\%$ relative to the baseline (25.7\% vs. 4.6\%), demonstrating that CPE is particularly critical for complex layouts where spatial ambiguity is most severe.

\begin{table}
  \centering
  \caption{Experiments on multilingual text-to-image generation.}
  \begin{tabular}{l|cc|cc}
    \toprule
    & \multicolumn{2}{c|}{Single-line} & \multicolumn{2}{c}{Multi-line} \\
    Method & NED $\uparrow$ & Sen. Acc $\uparrow$ & NED $\uparrow$ & Sen. Acc $\uparrow$ \\
    \midrule
    LoRA (w/o CPE) & 0.6326 & 0.3899 & 0.1924 & 0.0461 \\
    LoRA (w/ CPE) & \textbf{0.6765} & \textbf{0.4698} & \textbf{0.4813} & \textbf{0.2571} \\
    \bottomrule
  \end{tabular}
  \label{tab:char-pe}
\end{table}

\section{Qualitative Comparison of Subject Extension Mitigation}
\label{fig:controlnet}
\begin{figure*}[!b]
  \centering
  \includegraphics[width=\textwidth]{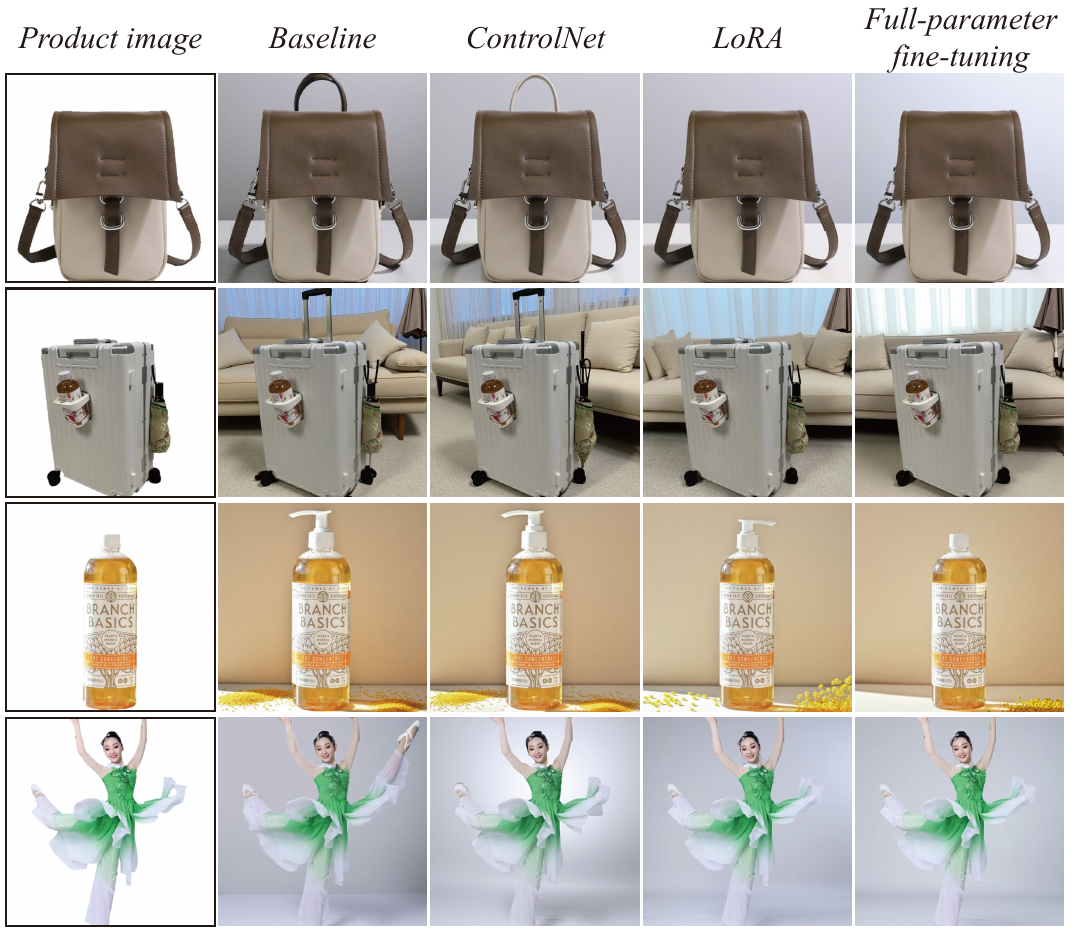}
  \caption{Samples with promotional text.}
  \label{control-vis}
\end{figure*}

Figure~\ref{control-vis} compares generation results across four configurations: (1) vanilla FLUX-Fill baseline, (2) ControlNet-augmented, (3) LoRA-tuned, and (4) our full-parameter fine-tuning approach.

\section{Ablation on Data Scale}
\label{supp:data-scale}

We conduct an ablation study to investigate how much data is required to eliminate subject extension. Following the protocol in Section 3 of the main paper, we fix the total training iterations while varying dataset size: 3K images for 300 epochs, 30K images for 30 epochs, and 300K images for 3 epochs.

As shown in Table~\ref{tab:data-scale}, surprisingly, even 3K training images reduce the extension rate from $41\%$ to $3.6\%$, outperforming ControlNet trained on 300K images ($23.6\%$). We further reduce training epochs for the 3K subset to 3 and 10 epochs; performance drops to $17.3\%$ and $9.3\%$, respectively, indicating that sufficient iterations are necessary for convergence.

\begin{table}[h]
  \centering
  \caption{Subject extension rate under varying data scales.}
  \begin{tabular}{l | c}
    \hline
    Method & Subject Extension Rate $\downarrow$ \\
    \hline
    Vanilla FLUX-Fill & $41.0\%$ \\
    ControlNet & $23.6\%$ \\
    Full tuning (3K, 3 ep) & $17.3\%$ \\
    Full tuning (3K, 10 ep) & $9.3\%$ \\
    Full tuning (3K, 300 ep) & $3.6\%$ \\
    Full tuning (30K, 30 ep) & $2.0\%$ \\
    Full tuning (300K, 3 ep) & $0.6\%$ \\
    \hline
  \end{tabular}
  \label{tab:data-scale}
\end{table}


\section{Automatic evaluation on subject preservation}
\label{supp:auto-eval}
Since the extension rate relies on human evaluation, we additionally provide an automatic evaluation method here. Specifically, we crop the subject regions, extract embedding using DINO v3~\cite{siméoni2025dinov3}, and compute the cosine similarity between the original product and the generated product to automatically measure subject fidelity. We report DINOv3~\cite{siméoni2025dinov3}. similarity on the segmented subject to quantify consistency Table~\ref{tab:dino-similarity}.  Model-based metrics are less reliable than human evaluation as they may miss fine-grained errors (e.g., subtle text artifacts, minor texture distortions). We recommend treating these as supplementary indicators only.

\begin{table}[h]
  \centering
  \caption{Subject similarity in DINOv3.}
  \begin{tabular}{l | c}
    \hline
    Method & DINOv3 similarity $\uparrow$ \\
    \hline
    FLUX Kontext(pro) & 0.9123 \\
    Step1x-Edit & 0.9050 \\
    Gemini2.5Flash & 0.9366 \\
    SeedEdit3 & 0.9465 \\
    PosterMaker & 0.9706 \\
    Ours & \textbf{0.9811} \\
    \hline
  \end{tabular}
  \label{tab:dino-similarity}
\end{table}

\section{Complete Prompts for Visualized Examples}
\label{supp:prompts}

All text prompts corresponding to visualized examples in the main paper are provided below. Original Chinese prompts are translated to English. For general editing models (SeedEdit~3/DreamPoster, Gemini~2.5 Flash, FLUX-Kontext, Step1x-Edit), we prepend the standardized prefix \texttt{``Background generation task.''} to all prompts.

\paragraph{Examples without promotional text.} See Figure~\ref{fig:cases-wo-text}.

\begin{figure*}[!t]
  \centering
    \includegraphics[width=0.8\textwidth]{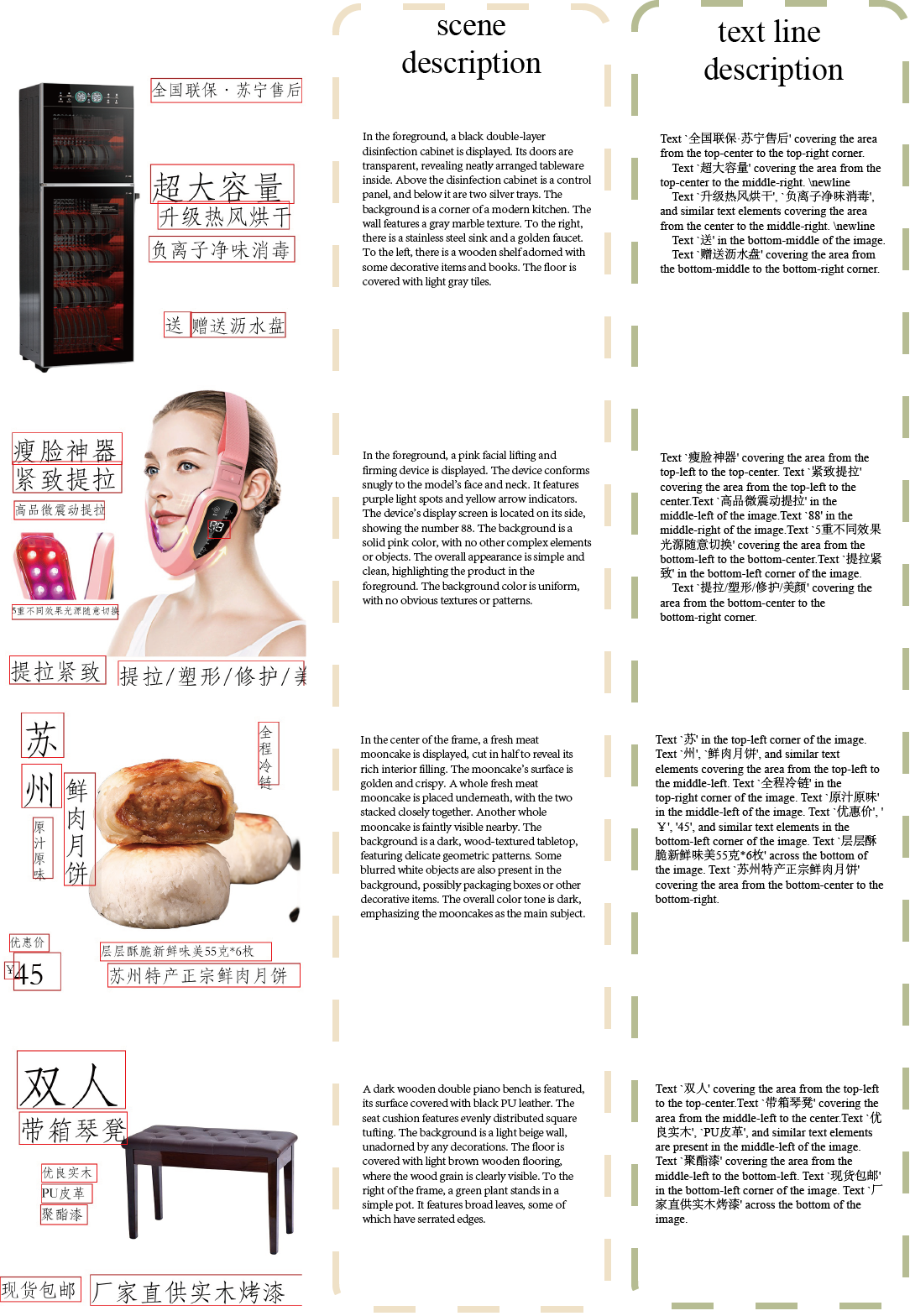}
\caption{Samples with promotional text. }
  \label{fig:cases-w-text}
\end{figure*}

\begin{figure*}[!t]
  \centering
  \includegraphics[width=\textwidth]{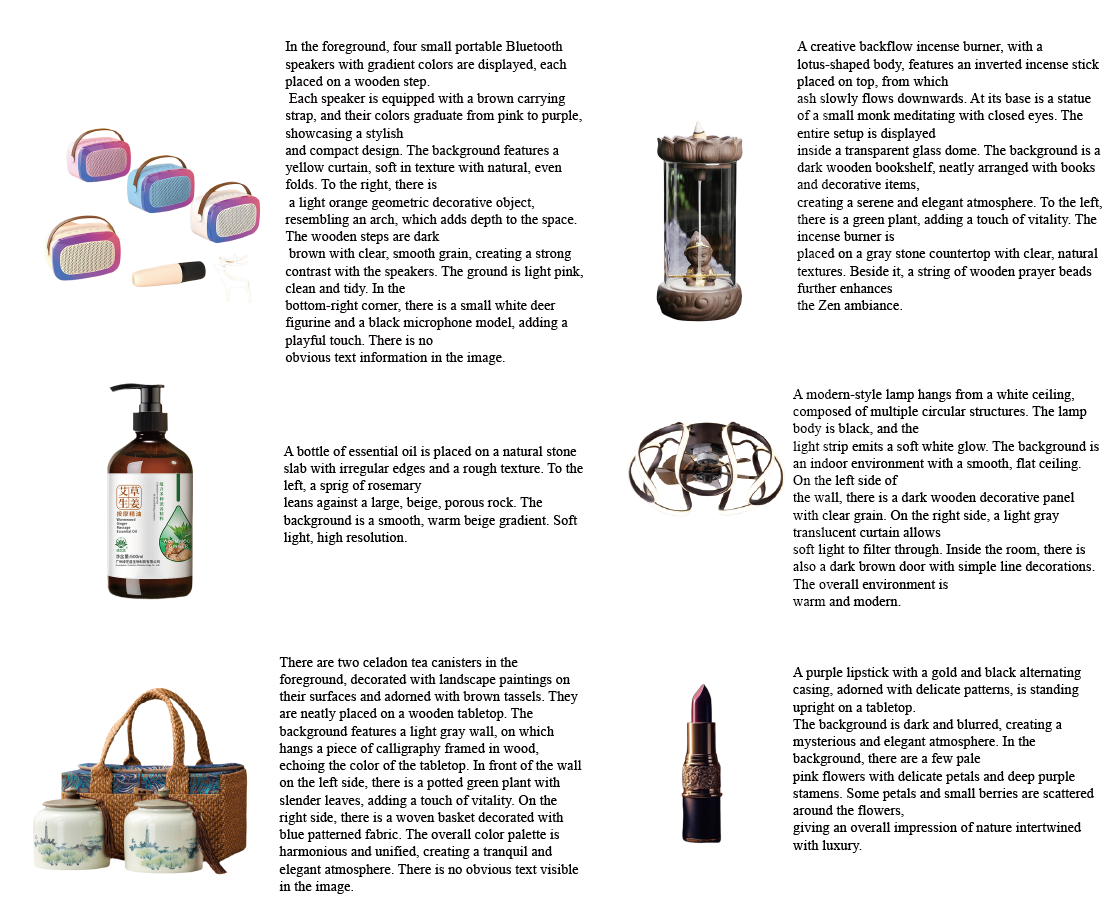}
  \caption{Samples without promotional text.}
  \label{fig:cases-wo-text}
\end{figure*}

\paragraph{Examples with promotional text.} See Figure~\ref{fig:cases-w-text}.


\clearpage
\newpage

{
    \small
    \bibliographystyle{ieeenat_fullname}
    \bibliography{main}
}

\end{document}